# Twitter Sentiment Analysis of Covid Vaccines

Wenbo Zhu and Tiechuan Hu

Courant Institute of Mathematical Sciences, New York University, New York, US
wz1305@nyu.edu
th2160@nyu.edu

## Abstract

*In this paper, we look at a database of tweets sorted by various keywords that could indicate the users' sentiment towards covid vaccines. With social media becoming such a prevalent source of opinion, sorting and ranking tweets that hold important information such as opinions on covid vaccines is of utmost importance. Two different ranking scales were used, and ranking a tweet in this way could represent the difference between an opinion being lost and an opinion being featured on the site, which affects the decisions and behavior of people, and why researchers were interested in it. Using natural language processing techniques, our aim is to determine and categorize opinions about covid vaccines with the highest accuracy possible.*

## Keywords

*Sentiment Analysis, Ranking algorithm, Machine learning classification*

## 1. Introduction

In the dataset that we created, there is a training and tokenized set of 3,000 tweets, each of them with different fields. Firstly, there is an id that identifies which tweet the algorithm is looking at. Then, there is a keyword that identifies which keywords caused the tweet to be flagged as a potential opinion about covid vaccines. This field will be very useful in determining which tweets are real opinions and which aren't, since some keywords have a much higher probability of flagging an opinion on covid vaccine than others.

The first step to create a model and the classifier is to understand how the data looks like. We will use several graph distribution graphics to understand our data in order to feed our model with the highest precision prediction number possible. For instance, we will calculate the number of characters per tweet in relation to positive and negative opinions and analyze the potential relationship between the two. We will also look at the stop words, and analyze if they are good predictors of whether an opinion is a stronger positive or stronger negative. For the classifier to work, we will need to make the text lowercase and have both punctuations and hashtags removed from the text. We intend to use a module called TweekTokenized to isolate punctuation that isn't desirable. We also hope to preprocess the text by removing emojis. There are a few very broad ranges of Unicode that is used for emojis on Twitter, and we will try to figure out an effective function to remove them. We believe that this would also be a useful step in preprocessing the data.

To create our prediction model we will use a training algorithm that will use the matrix created by the algorithms described to assign weights for every tweet. The ranking algorithm applied by the libraries of NLTK--VADER and TextBlob--and the range varies for these libraries. The libraries already have pre-trained models that give out the score. Once researchers input the sentences inside, the score would be outputted. The dataset is gained from Twitter API, Some keywords used are vaccine-related like vaccine, Pfizer, immunity and etc. The hypothesis is that the more interactions the tweets have, the more extreme, negative or positive, they are becoming.

## 2. Background

Twitter is the most popular technology where users are able to post, share, and make comments in the network space. Within the past years, Twitter gradually climbed on the top of social media letters worldwide. Since the popularity of social media skyrocketed, Twitter develops into the major method of communication for people. Like the slogan saying, "Twitter it's what's happening." As a matter of fact, many users use Twitter to unfold their own lives to the public. Therefore, tracking Twitter's data can provide a different perspective of what major events are occurring and help people understand others.

Besides Twitter, Covid 19, the widely-influenced pandemic, hits the world in a most unexpected way. Researchers, on the other hand, apply the natural language processing method to precisely discover the public's attitude towards the covid-19 vaccine. Along with the TWiki API, it allows us to access a huge amount of tweets history. The dataset is accurately based on the performance of the tweets' history. Since accessing API is an essential operation for data science, and researchers are able to use it as a tool to identify and analyze datasets, it pushes the natural language processing technique to higher ground. Meanwhile, researchers love using Twitter as the source for data because it is representative of social media and people's opinions.

## 3. Related Work

### 3. 1 Aspect-Based Sentiment Analysis

A challenge Dataset and Effective Models for Aspect-Based Sentiment Analysis proposed a new large-scaled Multi-Aspect Multi-Sentiment(MAMS) dataset that each sentence contains at least two aspects with different sentiment polarities, and compared with Aspect-based sentiment analysis (ABSA) techniques that only contains one or multiple aspects with the same sentiment polarities. In our project, we gather data from Twitter with Twitter API and use the sentiment analysis ranking algorithm to rank the tweets from -10 to 10 on people's sentiment towards covid vaccines. We do not have a MAMS dataset that has different polarities, but we do have a similar sentiment analysis ranking algorithm towards the sentiment polarity. Our group could compare our final sorted rankings and graph with the MAMS model.[1]

#### 3.1.1 Aspect Based Sentiment Analysis in Resource-Limited Languages

In Dataset Creation and Evaluation of Aspect Based Sentiment Analysis in Telugu, a Low Resource Language, researchers used ABSA in the Telugu dataset, analyzing the three tasks namely Aspect Term Extraction, Aspect Polarity Classification, and Aspect Categorization. It also developed a baseline system using deep learning methods to demonstrate its reliability. In our project, we are using English tweets for the analysis. The article is about the classification of the low resource language, Telugu, which is different from our research. The sentiment analysis of the Telugu is using extraction, classification, and categorization at the different aspects. It is not a low resource language, so we are allowed to process multiple tweets for dividing the corpus into training sets, development sets, and test sets. Putting a scale on the sentiment analysis for the polarity classification in our project seems to categorize the tweets in an efficient way. [3]

In the passage of Reliable Baselines for Sentiment Analysis in Resource-Limited Languages: The Serbian Movie Review Dataset, researchers present a dataset balancing algorithm that minimizes the sample selection bias by eliminating irrelevant systematic differences between the sentiment classes. It used the Serbian movie review dataset, SerbMR, to optimize the sentiment classification using machine learning features. In our project, our baselines for sentiment classification are based on the training sets of the dataset that is collected from the tweets on Twitter. In the paper, the Serbian movie review dataset is resource-limited and our dataset is not resource-limited. We can optimize and sort the sentiment classes based on their emotional scale from -10 to 10. Moreover, we interpret the output that was generated from the sentiment analysis ranking algorithm to make reasonable conclusions. [4]

### 3. 2 Self-reflective Sentiment Analysis

According to the self-reflective sentiment analysis, it is using machine learning classifiers based on n-grams, syntactic patterns, sentiment lexicon features, and distributed word embeddings with the novel mental health dataset. In our project, the machine learning classifiers can also be used in our sentiment analysis ranking algorithm. It is possible to apply n-grams, syntactic patterns, sentiment lexicon features, and distributed word embeddings in our algorithm. Also, we are using the Twitter dataset compared with the novel mental health dataset in this paper. The task definition of the machine learning classifiers might be the same as we are analyzing a similar type of dataset. Although our project is not self-reflective sentiment analysis, our project focus on the pessimistic or optimistic views of the tweets. [2]

### 3. 3 Evaluating Different Sentiment Analysis System

In Examining the Gender and Race Bias in Two Hundred Sentiment Analysis Systems, researchers used the Equity Evaluation Corpus (EEC) to examine 219 automatic sentiment analysis systems. The results show that several of the systems have significant bias, which provides higher sentiment intensity predictions for one race or one gender. In our project, we can examine the 219 automatic sentiment analysis systems, and see which one works best in our scenario. As we are only concerned with the sentiment analysis ranking algorithm, we are going to examine the EEC with the not-biased sentiment analysis systems. Also, we are going to make sure there is no significant bias or higher sentiment intensity predictions for certain races or gender in our system.[5]

## 4. Methodology

In the experiment, we have a database of tweets classified based on a keyword search that could indicate users' sentiment towards covid vaccines.

### 4. 1 Research question

Our simplest stage is that we are using the two libraries that already exist to do sentiment analysis. The dataset is gained by random search of covid-19 vaccine-related words like Pfizer, covid vaccine and etc. We already have two systems with the built-in library, and we refined the graph on the produced output. The research question is the hypothesis we are trying to prove, it is where the more interaction the tweets have, the more extreme sentiment will have.

### 4. 2 Dealing with the Outlier

Since a single follower could get a lot of followers which is equal to interactions in our experiment, it is considered as an outlier to the data. The people who got a lot of followers leads to skewed the data, thus we used a range to restrict the twitter participants who had 400-500 followers. Therefore, the participants in the data have the same follower base, so that the data does not have any outliers.

## 5. Experiments

The ranking algorithm derives from the difference between an opinion being lost and an opinion being featured on the site, which determines people's attitudes toward the vaccine. The sentiment scale for each of the automaton systems is different.

### 5. 1 Ranking algorithm

The first system used is the Vader sentiment analyzer that gives the compound score. It is accessible in NLTK and can directly be applied to the texts to detect the polarity of texts. Vadar contains a dictionary that each lexical word has its own corresponding emotion intensities. Summing up the whole sentence, Vader could detect the polarity of one tweet. The strength of Vader is that its analysis is accurate because it divides the sentences into different lexical categories and it has a solid rule to determine the polarity of text. Not only the Vader model would output negative, positive, and neural scores, but also you will get a compound score for the sentence or the whole document. It focuses on whether each sentence is positive or negative, which could further arrive at an overall opinion. For example, below, the VADAR system divides each lexical and computes the value for each one. Due to the positive output, it shows this sentence is positive, and the sentiment of the sentence corresponds with the library result.

**The numerical representation of the Vader library**

```
sentence = "I like polar bears"
sid.polarity_scores(sentence)

{'compound': 0.3612, 'neg': 0.0, 'neu': 0.444, 'pos': 0.556}
```

TextBlob is also another python library that processes texts. The API is simple and user-friendly. Another big advantage is it offers a lot of features like sentiment analysis, pos-tagging, and noun phrase extraction, retweet counts, and follower counts. TextBlob gives us one polarity sentiment score that determines the polarity and subjectivity of a sentence. The textblob sentiment score ranges from -1 to 1, which represents the negative and positive sentiment. Furthermore, the TextBlob has the capacity to reverse the negation word in order to acquire the correct polarity. The sentiment polarity was calculated through the intensity of the word, and if a sentence does not have any words that contain polarity, the TextBlob would return 0.0 as the result. It is a clear indication that both subjectivity and polarity tokens will result in a moderate or average sentiment score in a sentence. The varied polarities between different words in a sentence will be diffused throughout the integration of any words that had a polarity. Below are the results of the textblob. The textblob offers the retweet count and follower count for each random tweet, and it also calculates the sentiment. Research tests the reliability of TextBlob by inputting the sentence "this place is the worst", and the result is -1.0, corresponding to the actual sentiment score since the "worst" shows the extreme sentiment.

**The numerical representation of textblob library**

| | tweet | like_count | reply_count | retweet_count | retweeted | follower_count | sentiment |
|---|---|---|---|---|---|---|---|
| 1 | 3 Winnipeg communities to get priority vaccine... | NaN | NaN | 0 | NaN | 606.0 | 0.000000 |
| 2 | Justin Trudeau gets his first dose of the As... | NaN | NaN | 3 | NaN | 53.0 | 0.250000 |
| 3 | I took my dose on the heaviest day of my peri... | NaN | NaN | 0 | NaN | 148.0 | -0.750000 |
| 4 | 2nd shot in the arm\nMore emotional than I tho... | NaN | NaN | 0 | NaN | 831.0 | 0.166667 |
| 5 | It's time to break free of Government sponsor... | NaN | NaN | 20 | NaN | 697.0 | 0.075000 |

```
TextBlob('this place is the worst').sentiment.polarity

-1.0
```

Afterward, it will generate three CSV files, train.csv, test.csv, and dev.csv in our folder.
Two systems utilize different methods of training. Plotting all two of these systems on the graph, comparison with retweets account. If both systems show the correlation between interactions and sentiments simultaneously, it is more likely that this correlation is valid since they are all tested on the same data but trained with a different model. The accuracy might vary depending on which system is being applied, but if the correlation is the same through the two systems, it means that the hypothesis is strong.

### 5.2 Keyword classification

The keywords we used to build our dataset were among the most popular words spoken during the pandemic. The function of the keywords was to obtain covid vaccine-related tweets.
The polarity classification of the Vader allows us to detect if the text is expressing positive or negative opinions. The Vader model does not care if it is a subjective or objective sentence, it only takes account of the polarity of the words in a sentence. By searching covid vaccine-related tweets, the sentiment of related tweets will be analyzed through the Vader model in a way that certain words with polarity will be identified and taken into account.
The classification method used in TextBlob breaks the training sets into words and sentences, the Sentence objects function as classifying sentences. The classifiers module and API in the TextBlob pass the argument and document as its features in the extracting process. The Lemmatization of words categorizes each word in TextBlob.words and the lemmatize method can identify the part of speech in the WordNet library.

## 6. Results

Our evaluation plan is the graph analysis which represents the data, and its purpose is to check if there is consistency. If there is consistency, it means that the data is giving us relevant information that leads to the answer to our hypothesis.

### 6.1 Strategy for solving the problem

We have gathered and cleaned relevant data to answer our hypothesis of the research question. We scraped data from Twitter using their Twitter API, TWiki and stored the information in a file. Next, we used different sentiment analysis ranking algorithms to rank the relevant tweets on a scale from negative to positive sentiment towards covid vaccines. Next, these rankings will be organized to find relevant meaning. Hopefully, we can make a graph showing the large-scale sentiment towards vaccines on social media from the data we get from Twitter. In terms of what model we plan to use, we are keeping our options open. One that is a likely option is that of the VADER sentiment analyzer.
The problem with using NLTK libraries to analyze the sentiments is that if a Twitter account has thousands of followers, no matter what this account posts, he will gain enormous interactions, which excuse the date from what researchers attempt to find. Similarly, if one account has few followers, having few interactions but has a high sentiment score. Both cases are outliers, and to solve this problem, researchers take the follower account and divide it by retweet accounts, normalizing the range. The two libraries are Vader and textblob. The goal is to include all of them to exclude the potential outliers. Thus, the validity of the experiment is assured.

### 6.2 Data Analysis

The graph is achieved by sebron. The graphs are to describe the relationship between score and sentiment. The x-axis is the score while the y-axis is the sentiment score.

The first graph is the bar graph. It is counting the total number of negative, neutral, and positive tweets. The total tweets researchers obtained from Twitter API is 3000, reflected in this graph.

**The graphical representation of tweet counts and sentiment**

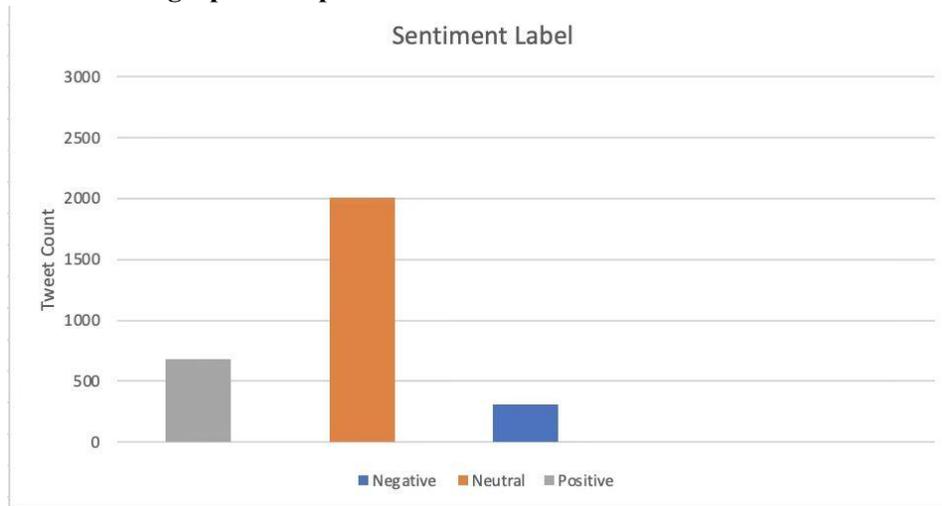

In the second graph, it is the plotting graph out of 3000 tweets. The sentiment is on the y-axis and the score is the x-axis. It is counting the total number of negative, neutral, and positive tweets. The orange is Vader, NLTK library, and the blue is the textblob. Researchers plot them on the same graph for a better comparison and contrast.

**The graphical representation of the Vader model**

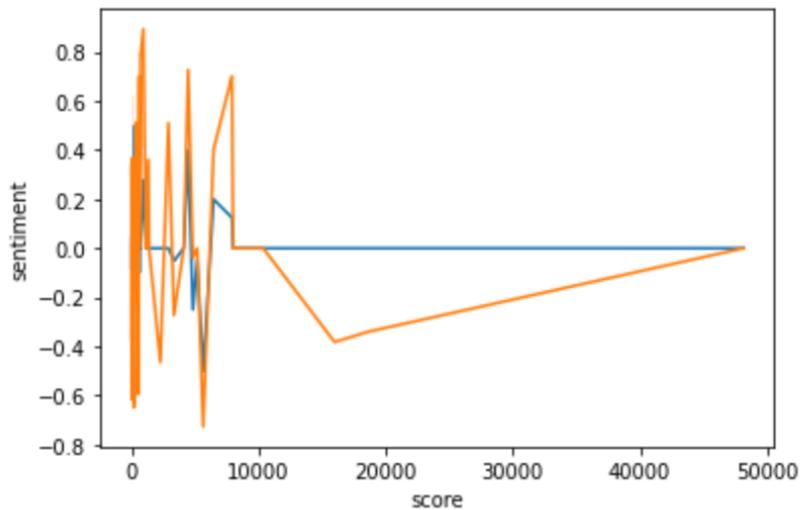

For the third graph, the sentiment is on the x-axis, and score as the y axis. Similar to the second graph, the orange is Vader, NLTK library, and the blue is the textblob. Researchers plot them on the same graph for a better comparison and contrast.

**The graphical representation of the Textblob model**

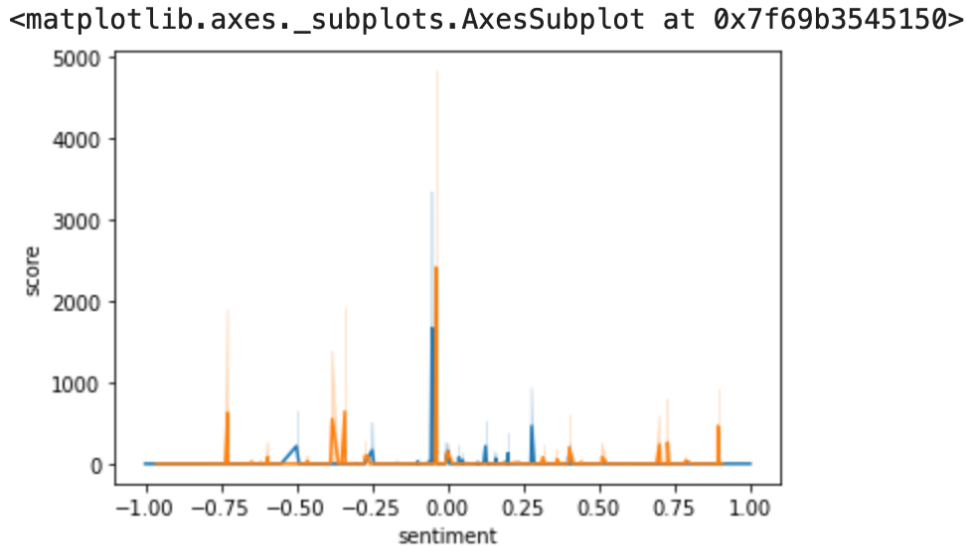

From the features of the three graphs, the two graphs out of three ( not counting the bar graph), the neutral and negative sentiments are getting the best results. For the second graph, the blue, textblob, is going for neutral all the way as the score increases. As for the Vader, there is a dip for the lower sentiment, but it still goes to neutral for the highest score. In the last graph, there is a spike when sentiment equals zero, which means at neutral, the score is the highest. The conclusion is that tweets, from neutral to negative, get the best interactions. However, there might be potential outliers. If a Twitter account has many followers, the score might skew the result. Consequently, the researchers only select accounts that have followers numbers ranging from 400 to 500.

## 7. Discussion

While we are working on the project, we find the sentiment analysis technique to be very effective at evaluating the public sentiment in the lexical approach. The conventional approach to capture dataset for sentiment analysis is derived from the TWiki API using different applications and libraries of the sentiment analysis technique. It is very quick to implement the technique because of its available libraries. Although the use of NLTK, TextBlob, VADER, and other machine learning classification approaches is very efficient at creating and evaluating the training, development, and evaluation dataset, it also has some disadvantages and weaknesses in it.

### 7.1 Weakness of Each Sentiment Analysis Technique

Each sentiment analysis has its own advantages and disadvantages, and we should implement the right ones and avoid certain try-except cases in our experiment.

The TextBlob is more often not able to handle complex or bizarre cases, for example, either if a sentence does not have words with specified polarity or TextBlob has an average sentiment score in the training set, the library would return 0.0 as the result.

Although the pre-trained VADER models are very good at identifying predetermined positive or negative scores, this lexical approach was based on the previous labeled sentiment analysis score from the training set. Therefore, the accuracy and reliability of the model are largely dependent on the training document.

### 7.2 General fallacy and disadvantages of the Sentiment Analysis Technique

There is the possibility that word spelling and grammatical mistakes cause different or wrong interpretations of the sentence. Moreover, the sarcasm and irony might not be easily identified and reviewed differently from what we expected. Certain jargon and memes are also not easy to be recognized, which greatly influences the performance of the system.

## 8. Conclusion

This research is to extract training and tokenized sets from 3000 tweets and categorize them into fields. The tokenized sets are implemented by VADER-NLTK and TextBlob, while the data visualization was achieved by Seabron. The conclusion is, neutral and negative sentiments are resulting in the most score. Thus, the neutral to negative retweets is the representation of people's sentiment. People hold neutral and negative attitudes towards covid-19 vaccines.

## 9. Future Work

The future work involves modifying our systems to generate better results: integrating the custom sentiment analysis into the NLTK library could provide a more accurate result. The further step is to combine the NLTK systems with more machine learning techniques, outputting more intelligent results. A further improvement is to explore richer linguistic analysis, for example, parsing, semantic analysis, and modeling. Finally, the researcher group could develop their own models since the reliability of the experiment is largely dependent on the training document, and creating researchers' own model would enhance the reliability.

## Acknowledgments

We would like to express our gratitude to every single author of the reference, this work is impossible to complete without your previous efforts. This article was completed under the guidance of Professor Adam Meyers, Mentor Vidit Vineet Bhargava. We appreciate the guidance in natural language processing from Professor Meyers and Mentor Bhargava, which broad our selection of content and writing specifications. At the same time, We thank Mentor Bhargava for supervising us and providing instructions on the paper. At last, We would like to thank everyone who has helped us with the paper, and your assistance has made this work possible and valuable.

**References**

1. Jiang, Q., Chen, L., Xu, R., Ao, X., & Yang, M. (n.d.). A challenge dataset and effective models for aspect-based sentiment analysis. Retrieved from https://www.aclweb.org/anthology/D19-1654/

2. Shickel, B., Heesacker, M., Benton, S., & Ebadi, A. (n.d.). Self-Reflective Sentiment Analysis. Retrieved from https://www.aclweb.org/anthology/W16-0303.pdf

3. Reddy, R. Y., Reddy, G. R., & Mamidi, R. (n.d.). Dataset Creation and Evaluation of Aspect Based Sentiment Analysis in Telugu, a Low Resource Language. Retrieved from https://www.aclweb.org/anthology/2020.lrec-1.617.pdf

4. Batanović, V., Nikolić, B., & Milosavljević, M. (n.d.). Reliable Baselines for Sentiment Analysis in Resource-Limited Languages: The Serbian Movie Review Dataset. Retrieved from https://www.aclweb.org/anthology/L16-1427.pdf

5. Kiritchenko, S., & Mohammad, S. (n.d.). Examining gender and race bias in two Hundred sentiment analysis systems. Retrieved April 25, 2021, from https://www.aclweb.org/anthology/S18-2005/

**Authors**

Wenbo Zhu and Tiechuan Hu, the authors of *Twitter Sentiment Analysis of Covid Vaccines*.

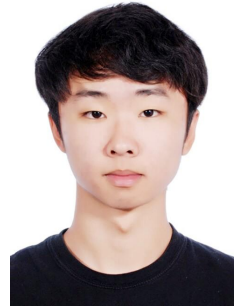

They are named on the dean's list as honor students this year in college for great academic achievement. Wenbo was born in Beijing, China. He started programming since 12, and holds a high passion for NLP. Tiechuan was born in Chengdu, China. He used to compete in AMC in high school, and shows a tremendous interest in NLP as well.

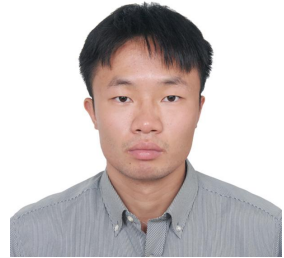